\documentclass{article}

\usepackage[nonatbib,preprint]{neurips_2024}

\usepackage[utf8]{inputenc} 
\usepackage[T1]{fontenc}    
\usepackage{hyperref}       
\usepackage{url}            
\usepackage{booktabs}       
\usepackage{amsfonts}       
\usepackage{nicefrac}       
\usepackage{microtype}      
\usepackage{xcolor}         
\usepackage{adjustbox}
\usepackage[style=numeric, sorting=none]{biblatex}
 \usepackage{multirow}
\usepackage{graphicx}

\title{From Bias to Balance: Detecting Facial Expression Recognition Biases in Large Multimodal Foundation Models}

\author{%
  Kaylee Chhua\thanks{Lead author.} \\
  Lake Wales High School \\
  Lake Wales, FL 33853 \\
  \texttt{kayleechhua@gmail.com} \\
   \And
  Zhoujinyi Wen\thanks{Equal contribution.} \\
  Thomas Jefferson Sr. High School \\
  Bloomington, MN 55437 \\
  \texttt{zhjnyi.wen@gmail.com} \\
   \AND
  Vedant Hathalia\footnotemark[2]\\
  Bellarmine College Preparatory\\
  San Jose, CA 95126 \\
  \texttt{vedanthathalia@gmail.com} \\
  \And
  Kevin Zhu \thanks{Senior Author}\\
  Algoverse AI Research\\
  Palo Alto, CA 94306 \\
  \texttt{kevin@algoverse.us} \\
  \And
  Sean O'Brien\footnotemark[3] \\\
  Algoverse AI Research\\
  Palo Alto, CA 94306 \\
  \texttt{sean@algoverse.us} \\
}

\begin{document}

\maketitle

\begin{abstract}
This study addresses the racial biases in facial expression recognition (FER) systems within Large Multimodal Foundation Models (LMFMs). Despite advances in deep learning and the availability of diverse datasets, FER systems often exhibit higher error rates for individuals with darker skin tones. Existing research predominantly focuses on traditional FER models (CNNs, RNNs, ViTs), leaving a gap in understanding racial biases in LMFMs. We benchmark four leading LMFMs: GPT-4o, PaliGemma, Gemini, and CLIP to assess their performance in facial emotion detection across different racial demographics. A linear classifier trained on CLIP embeddings obtains accuracies of 95.9\% for RADIATE, 90.3\% for Tarr, and 99.5\% for Chicago Face. Furthermore, we identify that Anger is misclassified as Disgust 2.1 times more often in Black Females than White Females. This study highlights the need for fairer FER systems and establishes a foundation for developing unbiased, accurate FER technologies. Visit \href{https://kvjvhub.github.io/FERRacialBias/}{https://kvjvhub.github.io/FERRacialBias/} for further information regarding the biases within facial expression recognition.

\end{abstract}

\section{Introduction}
Over the past decade, developments in facial expression recognition (FER), such as the introduction of deep learning algorithms \cite{onyema2021enhancement} and the creation of more diverse, extensive datasets \cite{huang2008labeled}, have accelerated its use in real-world scenarios \cite{dhall2015video}. FER is used to identify the emotional and physical well-being of blue-collar workers \cite{patwardhan2016emofitaffectmonitoringsedentary}, detect pain in patients who face difficulties with communication \cite{deuschel2020uncovering}, and assess a job candidate's facial expressions to score them on factors such as intelligence, personality, and criminality \cite{hirevue}. Despite their potential, these systems face a great challenge: pervasive racial bias \cite{fan2023addressingracialbiasfacial}. Numerous studies have highlighted that facial recognition technology often displays higher error rates for individuals with darker skin tones compared to those with lighter skin \cite{klare2012face, article, buolamwini2018gender}. Failure to mitigate racial biases in FER systems not only poses legal risks but perpetuates systemic inequalities. In the hiring process, FER is used to select the most “appropriate” candidates. However, such algorithms have been proven to be biased, placing minorities, such as women and people of color, in occupations characterized as “low competency” \cite{he2019stereotypes}. They have grown exponentially, with 98\% of Fortune 500 companies adopting them, causing a 270\% increase in usage over the short time span of four years \cite{tilmes2022disability}.

While racial bias in FER has been extensively studied, little research \cite{kopalidis2024advances}, \cite{sajjad2023comprehensive} has been conducted on the racial bias of Large Multimodal Foundation Models (LMFMs): an emerging, versatile technology capable of processing data from multiple modes such as text \cite{wei2022chain}, image \cite{li2023blip2}, \cite{alayrac2022flamingo}, and audio \cite{su2023pandagpt} with high accuracy \cite{li2024multimodal}. To address this issue, we (i) provide a comprehensive benchmark of four leading LMFMs to assess their performance in facial emotion detection across different racial groups, (ii) demonstrate the efficacy of self-supervised learning in mitigating the biases of LMFMs, and (iii) conduct statistical inferences such as two-proportion z-tests to calculate misclassification rates and identify potential biases between various races.

\section{Related Works}
\textbf{Deep Learning Architectures:} In recent years, researchers have become interested in Convolutional Neural Networks (CNNs) and hybrid models such as CNN-LSTM (Long Short-Term Memory) for facial emotion recognition \cite{mellouk2020facial}. These architectures are effective at extracting features from facial images, but often require large amounts of labeled data for training and suffer from overfitting when dealing with limited datasets \cite{patel2020facial}. Although LMFMs show potential due to their vast training networks \cite{yitzhak2024retrieved}, we discover that they struggle to capture subtle nuances in facial expressions.

\textbf{Dataset Selections:} Various factors such as illumination, noise, and blur can impair FER performance \cite{dutta2012impact}. Additionally, upscaling FER datasets to 224 x 224 pixels for neural networks often results in detail loss and reduced classification accuracy \cite{gomez2023facial}. To ensure a fair evaluation of Large Multimodal Foundation Models, we employ high-resolution and uniform datasets.

\section{Dataset Modifications}
When selecting datasets for benchmarking and fine-tuning our LMFMs, we prioritize those with both emotion and race labels. This allows us to assess model accuracy and identify racial biases by comparing the dataset’s ground truth emotions with their predicted emotions.

\textbf{RADIATE:} The RADIATE Dataset includes 1,721 1500 x 1500 96 DPI images portraying Anger, Calm, Disgust, Fear, Happy, Neutral, Sad and Surprise expressions across adults from Black, White, Hispanic, and Asian backgrounds \cite{conley2018racially}. We exclude the Other racial category due to insufficient data points. We remove the Calm category because after inspecting the images, Calm and Neutral show no distinct differences.

\textbf{Tarr:} The Tarr Dataset contains 1,226 250 x 250 72 DPI images of Black, White, Asian, and Hispanic adults with the expressions Anger, Confusion, Disgust, Fear, Happy, Neutral, Sad, and Surprise \cite{stimulusimages}. We exclude the Multiracial category due to a limited number of data points. Furthermore, the broad diversity within the Multiracial category leads to challenges in identifying specific biases. We also remove the Confusion category in Tarr due to a lack of data points in racial groups besides White.

\textbf{Chicago Face:} The Chicago Face Dataset comprises of 1,208 2444 x 1718 240 DPI images of Asian, Black, Latino, and White adults with the facial expressions Neutral, Happy, Anger, and Fear \cite{Ma2015TheCF}. The dataset mainly contains Neutral expressions, with 100\% of Asian and Hispanic data points displaying this emotion. After post-processing, we standardize all images to 224 x 224 pixels at 96 DPI and crop the Chicago Face Dataset images to ensure a square, centered composition.

\begin{figure}[hbt!]
    \centering
    \includegraphics[width=0.75\linewidth]{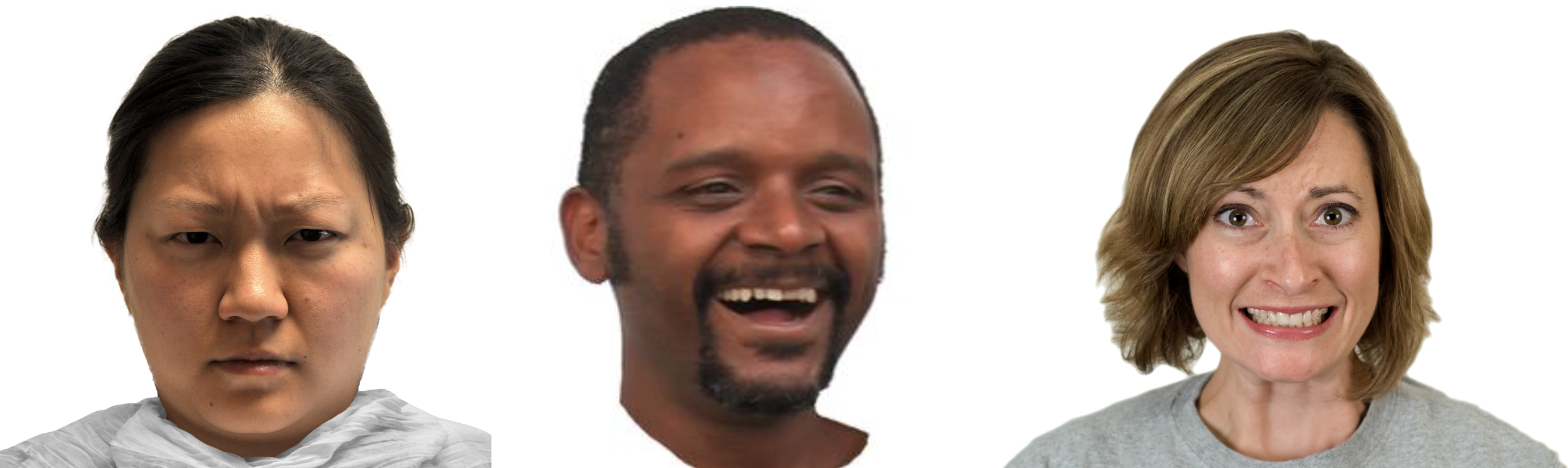}
    \caption{RADIATE (Left), Tarr (Middle), and Chicago Face (Right) Dataset Images}
\end{figure}

\section{Benchmarking Large Multimodal Foundation Models}

To investigate existing biases in current publicly available Large Multimodal Foundation Models, we zero-shot prompt four major models: GPT-4o, PaliGemma, Gemini, and CLIP. We choose these models because of their established reputation, widespread public usage, and foundational roles for various fine-tuned LMFM models that attempt to accomplish similar tasks. In this process, we observe and record the predicted emotions generated by each model. We then compare them against human-labeled emotions from the dataset to analyze accuracy across different races, expressions, and potential trends.

\textbf{GPT-4o and Gemini:} We prompt each image with: "Classify the image into one of the eight expressions: [Happy, Anger, Sad, Neutral, Calm, Disgust, Fear, Surprise]." To mitigate hallucinations, GPT-4o (gpt-4o-2024-05-13), Gemini 1.5 Flash (gemini-1.5-flash-001) and Gemini 1.5 Pro (gemini-1.5-pro-001) attempt classification up to ten times until a valid response is produced. We use Chain of Thought (CoT) zero-shot prompting, maj@5 sampling, and a temperature setting of 1 to yield optimal results.

\textbf{PaliGemma:} To benchmark PaliGemma, we use PaliGemma pt-224 (paligemma-3b-pt-224), PaliGemma mix-224 (paligemma-3b-mix-224), and PaliGemma pt-448 (paligemma-3b-pt-448) for zero-shot image classification. To prevent hallucinations, we adjust prompts to lowercase and use the tag 'answer en.' 

\begin{figure}
    \centering
    \includegraphics[width=1\linewidth]{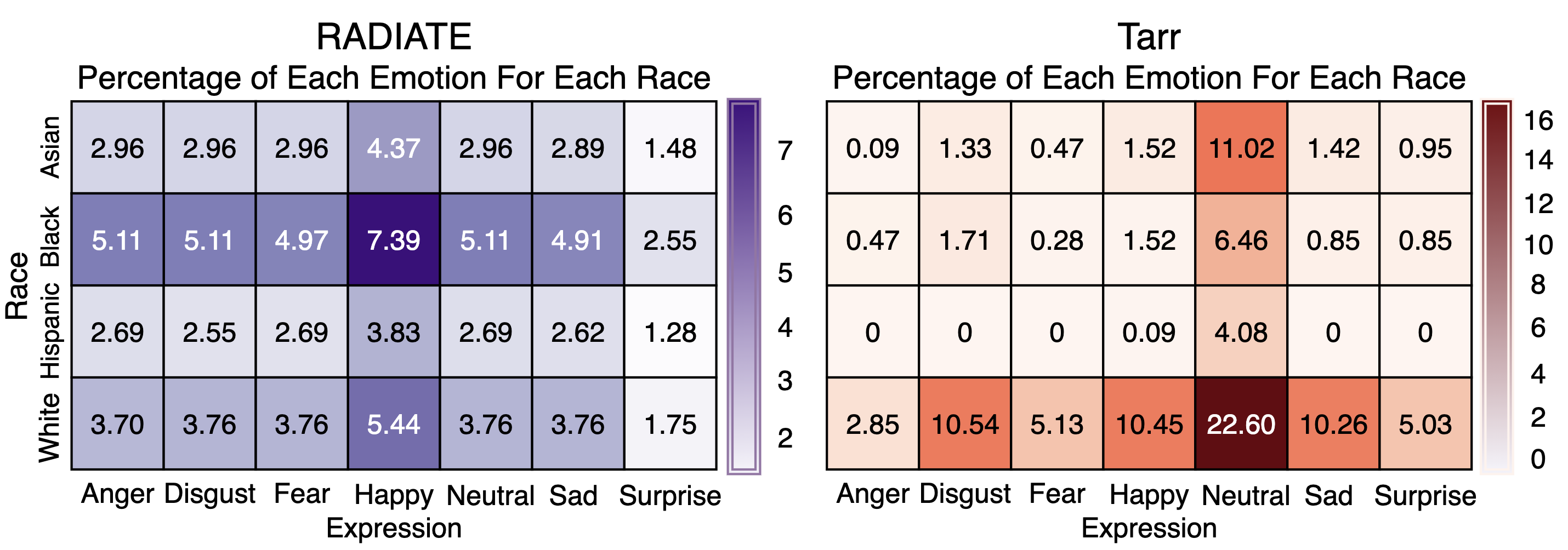}
    \caption{RADIATE and Tarr Race and Emotion Distribution}
\end{figure}

\textbf{CLIP:} We perform zero-shot classification using self-supervised learning by applying a logistic regression classifier on top of CLIP embeddings. These embeddings, along with numerical labels for expressions, train the logistic regression classifier. We then calculate the classifier’s accuracy and decode the model’s predictions back to emotion labels.

\section{Results and Discussion}

In evaluating the zero-shot prompting accuracy across the three datasets, we observe significant performance variations among the leading LMFMs in FER. The linear classifier on CLIP embeddings has the highest accuracy, achieving accuracies of 95.9\%, 90.3\% and 99.5\% on RADIATE, Tarr, and Chicago Face respectively. GPT-4o and GPT-4o with Chain-of-Thought (CoT) also demonstrate strong performance. Interestingly, GPT-4o did not consistently perform better when evaluated with maj@5 statistics. Conversely, PaliGemma mix-224, the best performing model of PaliGemma, demonstrates comparatively lower accuracy, suggesting limitations in its effectiveness for FER (see Table 1).

{\renewcommand{\arraystretch}{1.25}%
\begin{table}[hbt!]
    \centering
    \caption{Model (Temperature 1) and Method Comparison for Zero-shot Prompting Dataset Accuracy}
        \begin{tabular}{ccccccc}
        Model and Zero-Shot & \multicolumn{2}{c}{RADIATE} & \multicolumn{2}{c}{Tarr} & \multicolumn{2}{c}{Chicago Face} \\ \cline{2-7} 
        Prompting Method    & single        & maj@5       & single      & maj@5      & single           & maj@5         \\ \hline
        GPT-4o  & 70.9       & 70.5       & 71.5     & \textbf{71.7}      & 95.6          & 95.4         \\
        GPT-4o with CoT & 70.5       & \textbf{71.0}       & 70.0     & 70.8      & 94.7          & \textbf{95.9}         \\
        PaliGemma mix-224   & 45.7          & N/A       & 43.3        & N/A      & 73.7             & N/A         \\
        Gemini 1.5 Flash    & 67.2          & N/A      & 68.3        & N/A      & 92.9           & N/A         \\
        Gemini 1.5 Pro  & 70.9          & N/A       & 68.0       & N/A     & 94.5             & N/A          \\
        Linear Classifier on CLIP Embeddings      & \textbf{95.9}          & N/A      & \textbf{90.3}        & N/A    & \textbf{99.5}             & N/A         
        \end{tabular}
\end{table}
}

Our investigation into the performance of GPT-4o, Gemini 1.5 Pro, and CLIP in facial emotion recognition (FER) reveals distinct trends across different racial groups and emotions (see Table 2). CLIP demonstrates superior accuracy across all racial categories ranging from 94.2\% to 97.4\%, compared to GPT-4o (76.2\% to 84.5\%) and Gemini 1.5 Pro (74.1\% to 83.4\%).

Regarding emotions, CLIP consistently excels in recognizing emotions such as Happy and Neutral, achieving near-perfect scores with percentages ranging from 98.1\% to 100\%. However, CLIP misclassifies the emotions of Black Females in RADIATE and Tarr 2.1 times more often than its next highest misclassified demographic, Asian Males. GPT-4o and Gemini 1.5 Pro show varied performance across racial categories and emotions. For instance, Fear and Sadness prove to be challenging, particularly among Asian and White subjects, with recognition rates dropping as low as 16.3\% for Fear and 27.6\% for Sadness. In contrast, emotions like Happy and Surprise exhibit higher recognition rates across all models and racial groups, highlighting their relatively straightforward identification compared to more nuanced emotions.

{\renewcommand{\arraystretch}{1.25}%
\begin{table}[h]
    \centering
    \caption{GPT-4o, Gemini 1.5 Pro, and CLIP Overall Accuracy By Emotion and Race}
    \begin{adjustbox}{max width=1\textwidth}
        \begin{tabular}{ccccccccccccc}
        \multicolumn{1}{l}{} & \multicolumn{4}{c}{GPT-4o}                                                                                       & \multicolumn{4}{c}{Gemini 1.5 Pro}                                                                               & \multicolumn{4}{c}{CLIP}                                                                                         \\ \cline{2-13} 
        \multicolumn{1}{l}{}                  & \multicolumn{1}{l}{Asian} & \multicolumn{1}{l}{Black} & \multicolumn{1}{l}{Hispanic} & \multicolumn{1}{l}{White} & \multicolumn{1}{l}{Asian} & \multicolumn{1}{l}{Black} & \multicolumn{1}{l}{Hispanic} & \multicolumn{1}{l}{White} & \multicolumn{1}{l}{Asian} & \multicolumn{1}{l}{Black} & \multicolumn{1}{l}{Hispanic} & \multicolumn{1}{l}{White} \\ \hline
        Anger                                 & 57.8                      & 77.9                      & 70                           & 68.2                      & 44.4                      & 73.6                      & 60                           & 56.1                      & 91.1                      & 96.9                      & 92.5                         & 90.4                      \\
        Disgust                               & 72.4                      & 89.4                      & 97.4                         & 85.0                      & 81.0                      & 83.0                      & 92.1                         & 73.1                      & 84.5                      & 88.3                      & 89.5                         & 95.8                      \\
        Fear                                  & 16.3                      & 57.5                      & 40.0                         & 44.9                      & 36.7                      & 63.1                      & 47.5                         & 49.4                      & 91.8                      & 95.0                      & 100                          & 85.8                      \\
        Happy                                 & 88.9                      & 87.9                      & 87.9                         & 92.0                      & 91.4                      & 87.6                      & 91.4                         & 93.1                      & 100                       & 99.3                      & 100                          & 99.7                      \\
        Neutral                               & 93.7                      & 95.9                      & 97.4                         & 93.9                      & 97.4                      & 96.5                      & 99.0                         & 95.6                      & 100                       & 99.7                      & 99.5                         & 98.1                      \\
        Sad                                   & 46.6                      & 36.7                      & 56.4                         & 43.9                      & 27.6                      & 30.5                      & 46.2                         & 32.3                      & 82.8                      & 90.2                      & 92.3                         & 88.4                      \\
        Surprise                              & 75.0                      & 83.0                      & 100                          & 55.7                      & 75.0                      & 78.7                      & 94.7                         & 43.8                      & 93.8                      & 89.4                      & 100                          & 82.3                      \\ \hline
        Accuracy                              & 76.2                      & 81.1                      & 84.5                         & 77.1                      & 77.9                      & 80.2                      & 83.4                         & 74.1                      & 95.1                      & 96.6                      & 97.4                         & 94.2                     
        \end{tabular}
    \end{adjustbox}
\end{table}
}

Through the two-proportion z-tests done on misclassification by race, we find that the p-values for Asian vs. Black (0.024), Asian vs. Hispanic (0.011), Black vs. White (0.010), and Hispanic vs. White (0.007) are all statistically significant at 95\% confidence. Specifically, the z-score values for Asian vs. Black (2.25) and Asian vs. Hispanic (2.55) are positive, meaning the latter is more misclassified, while in Black vs. White (-2.57) and Hispanic vs. White (-2.70), the former is more misclassified. For zero-shot PaliGemma, Neutral is mistaken as Sad for White Females 3.38 times more often than their Black Female counterparts across all datasets. On the other hand, Fear is misclassified as Surprise 2.46 times and Anger is misclassified as Disgust 2.1 times more often for Black Females than White Females for zero-shot GPT-4o on the RADIATE dataset. Due to the overrepresentation of White images, which are more frequently misclassified, no other significant statistical trends or patterns were observed.

\section{Conclusion and Outlook}
We benchmark four leading Large Multimodal Foundation Models, GPT-4o, PaliGemma, Gemini, and CLIP, on three diverse datasets to assess their racial biases in facial expression recognition systems. Our findings demonstrate significant disparities in accuracy across different racial categories. We observe the highest misclassification rates in Black Females across all benchmarked models and discover lower accuracy rates in Sad and Fear among all races. Fine-tuning facial emotion recognition algorithms across races to reduce bias is crucial for applications in security, healthcare, and social interaction analysis \cite{mehta2019recognition, TheRegulatoryReview_2021, yolcu2019facial}. To improve social outcomes, future research must prioritize the development of adaptive AI systems that account for cultural nuances in facial expressions.

\printbibliography

\end{document}